%% file: main.tex
\documentclass[10pt,twocolumn,letterpaper]{article}

 \usepackage{cvpr}              


\definecolor{cvprblue}{rgb}{0.21,0.49,0.74}
\usepackage[pagebackref,breaklinks,colorlinks,allcolors=cvprblue]{hyperref}

\usepackage{multirow}

\def\paperID{6588} 
\def\confName{CVPR}
\def\confYear{2025}

\title{DiffPortrait360: Consistent Portrait Diffusion for 360 View Synthesis}

\author{Yuming Gu\textsuperscript{1,2}, Phong Tran\textsuperscript{2}, Yujian Zheng\textsuperscript{2}, Hongyi Xu\textsuperscript{3}, Heyuan Li\textsuperscript{4}, 
Adilbek Karmanov\textsuperscript{2}, Hao Li\textsuperscript{2,5}\\
\textsuperscript{1}University of Southern California, \textsuperscript{2}MBZUAI, \textsuperscript{3}ByteDance Inc.
\\
\textsuperscript{4}The Chinese University of Hong Kong, Shenzhen, \textsuperscript{5}Pinscreen Inc. \\ 
\href{https://freedomgu.github.io/DiffPortrait360}{https://freedomgu.github.io/DiffPortrait360}\\
{\tt\small yuminggu@usc.edu}, {\tt\small \{the.tran, yujian.zheng, adilbek.karmanov\}@mbzuai.ac.ae}\\
{\tt\small \ heyuanli@link.cuhk.edu.cn, hao@hao-li.com}
}

\begin{document}
\linespread{1.0}
\input{figures/teaser}

\input{sec/0_abstract}    
\input{sec/1_intro_hao}
\input{sec/2_Related_Work}
\input{sec/3_Method}
\input{sec/4_Experiment}

\input{sec/5_Conclusion}

{
    \small
    \bibliographystyle{ieeenat_fullname}
    \bibliography{main}
}


\input{sec/X_suppl}

\end{document}

%% file: figures/teaser.tex
\twocolumn[{%
\renewcommand\twocolumn[1][]{#1}%
\maketitle
\begin{center}
    \centering
    \captionsetup{type=figure}
    \includegraphics[width=\linewidth]{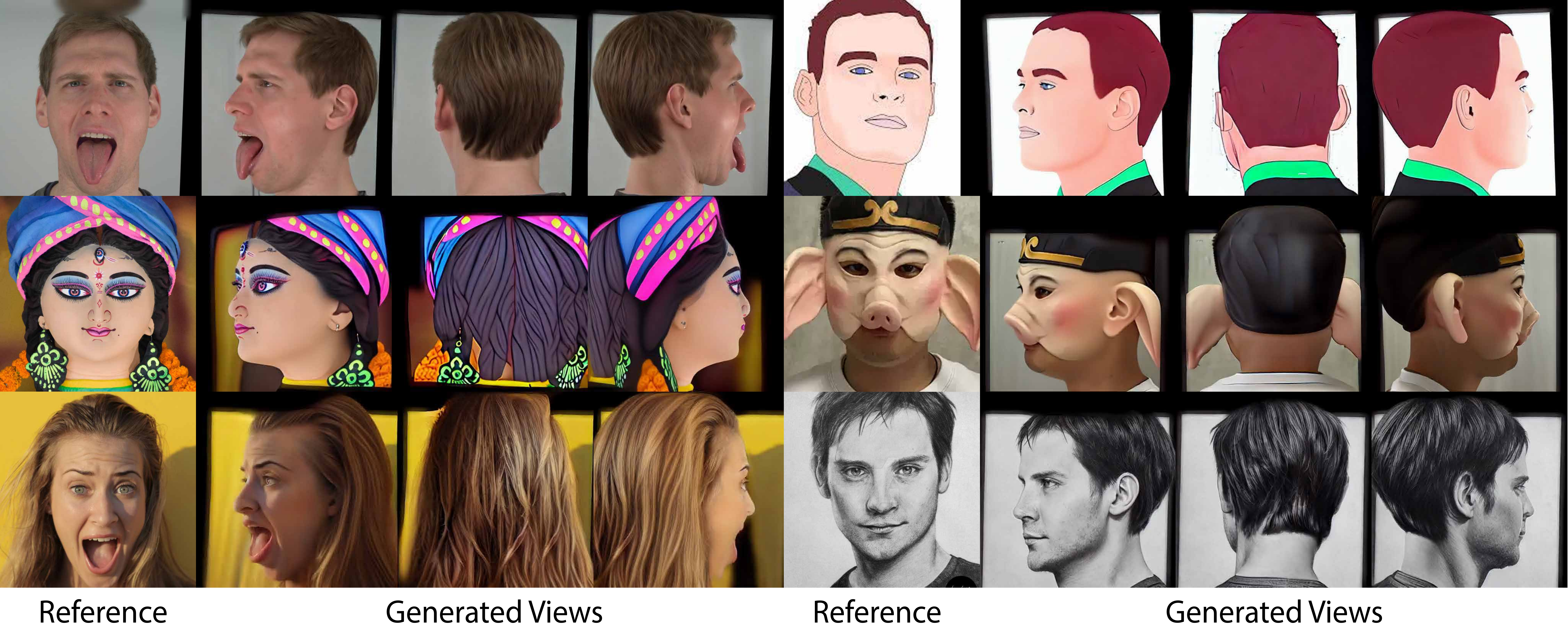}
    \captionof{figure}{Our DiffPortrait360 enables 360$^\circ$ view-consistent full-head image synthesis. It is universally effective across a diverse range of facial portraits, allowing for the creation of 3D-aware head portraits from single-view images.} 
    \label{fig:teaser}
\end{center}%
}]

%% file: sec/0_abstract.tex
\begin{abstract}
Generating high-quality 360-degree views of human heads from single-view images is essential for enabling accessible immersive telepresence applications and scalable personalized content creation.
While cutting-edge methods for full head generation are limited to modeling realistic human heads, the latest diffusion-based approaches for style-omniscient head synthesis can produce only frontal views and struggle with view consistency, preventing their conversion into true 3D models for rendering from arbitrary angles.
We introduce a novel approach that generates fully consistent 360-degree head views, accommodating human, stylized, and anthropomorphic forms, including accessories like glasses and hats. Our method builds on the DiffPortrait3D framework, incorporating a custom ControlNet for back-of-head detail generation and a dual appearance module to ensure global front-back consistency. By training on continuous view sequences and integrating a back reference image, our approach achieves robust, locally continuous view synthesis. Our model can be used to produce high-quality neural radiance fields (NeRFs) for real-time, free-viewpoint rendering, outperforming state-of-the-art methods in object synthesis and 360-degree head generation for very challenging input portraits.
\end{abstract}

%% file: sec/1_intro_hao.tex
\vspace{-4mm}
\section{Introduction}
\label{sec:intro}

In game, film, and animation production, CG characters are central to the content creation pipeline, meticulously designed with the right balance of realism and stylization to align with the narrative.
Multi-view stereo systems and 3D scanners are commonly used to streamline the tedious manual process of creating realistic human characters. However, stylized characters are still often 3D modeled from scratch. Alongside the growing demand for efficient and scalable 3D content production pipelines, immersive telepresence and virtual world applications are fueling interest in embodying personalized or imaginary characters created directly by end-users. In this context, highly accessible 3D modeling tools that can generate characters from a single photograph or drawing are becoming increasingly relevant.

The most widely adopted approaches for image-based full 3D head reconstruction combine parametric models, like 3D Morphable Models (3DMMs)~\cite{paysan20093d, 10.1145/3130800.31310887, wu2019mvf, dib2021high, tsutsui2022novel, zhuang2022mofanerf, DBLP:journals/corr/abs-2106-11423,pinscreenavatar,itseez3d}, with data-driven hair modeling or generate a single textured 3D mesh that includes both the face and hair. These methods often struggle with diverse hairstyles and capturing detailed appearance textures for the back of the head. 
Cutting-edge 3D head reconstruction, utilizing transformer-based networks~\cite{trevithick2023,tran2023voodoo,tran2024voodoo,deng2024portrait4dv2,deng2024portrait4d} to estimate tri-plane representation, have shown success in capturing complex portraits of human heads with diverse hairstyles and accessories. However, tri-plane introduces strong projection ambiguity for extreme non-frontal poses and therefore fails to model 360-degree head. To facilitate the novel synthesis of full 360-degree views, tri-grid~\cite{An_2023_CVPR} neural volume representations have been introduced. While these methods have demonstrated impressive results, they are limited to realistic human heads and cannot handle stylized or other anthropomorphic shapes.

The latest diffusion-based image-to-3D generation methods~\cite{wu2024unique3d,liu2023zero1to3,shi2023zero123plus} are designed to model a wide range of 3D objects. However, they struggle to capture detailed, domain-specific appearances, such as those of human heads and are prone to severe, unpredictable artifacts. A recent diffusion model, DiffPortrait3D~\cite{Gu_2024_CVPR}, demonstrates the capability to generate novel views of style-omniscient heads by incorporating 3D-aware noise for inference, and fine-tuning the model on a large 3D head dataset. In addition to being limited to frontal views, this approach also displays noticeable inconsistencies between angles.

We present the first method capable of generating \textit{consistent 360-degree} views from a single portrait of a head, accommodating human, stylized, or animal anthropomorphic forms, as well as accessories like glasses jewelries, or masks. Our novel views can be computed offline (5.6 sec per view) and transformed into a high-quality neural radiance field (NeRF)~\cite{mildenhall2020nerf}, enabling free-viewpoint rendering in real-time. Our method is particularly robust, capable of handling a wide range of subjects, complex hairstyles, varied head poses, and expressive facial features, including detailed elements like tongues.

Building on the state-of-the-art DiffPortrait3D framework, we introduce key innovations to support back-view renderings and achieve locally continuous view consistency. We begin by generating the back of the head from the input image using a custom ControlNet~\cite{zhang2023adding}, trained specifically to infer back-of-head details from a front-facing image. This model is trained on a carefully curated dataset featuring diverse, realistic, and stylized heads. The front and back images are then passed into a dual appearance module (stable diffusion UNet), which extracts appearance information from both perspectives to condition our main image generator (stable diffusion denoising network) through attention layers. Unlike DiffPortrait3D, our approach incorporates an additional back reference image to improve global appearance consistency. Furthermore, we ensure locally continuous view consistency by using \textit{continuous view sequences (8 views)} as conditions during training.
Unlike DiffPortrait3D, which is trained only on frontal 3D-aware head generations, our approach utilizes a dataset comprising full 360-degree views of human heads, with data that is both captured and synthetically generated. Finally, our denoising process also uses a 3D-aware noise which is initialized using a fine-tuned inversion process-based a StyleGAN-based 360 head generator \cite{An_2023_CVPR}.

We demonstrate that diffusion models can generate highly detailed and consistent 360-degree head views from in-the-wild images across an extensive range of styles and complex portraits, handling challenging lighting conditions, intricate hairstyles, diverse appearances, and varying head poses. Our approach significantly outperforms all the latest generative methods for our challenge human head benchmarks generation, w.r.t. key metrics. More importantly, our method produces globally consistent appearances for the back of the head and locally view consistent synthesis when generating 360-degree views, which enables us to reconstruct true 3D representations (e.g., NeRFs) for free-view point rendering.
We present the following contributions:

\begin{itemize}
\item We present a novel method for $360^\circ$ view synthesis from a single style-omniscient portrait, which leverages priors from pre-trained stable diffusion models to generate view-consistent content.

\item We propose a dual appearance control module, supported by a diffusion-based back-view generator, that ensures global consistency in novel view synthesis around the subject.

\item We introduce a novel training strategy that employs view-consistent sequence generation to enhance continuity and smoothness, utilizing a 3D head dataset composed of both captured and synthetic subjects.

\end{itemize}

\input{figures/pipeline}

%% file: figures/pipeline.tex
\begin{figure*}
\centering
\includegraphics[width=0.9\linewidth]{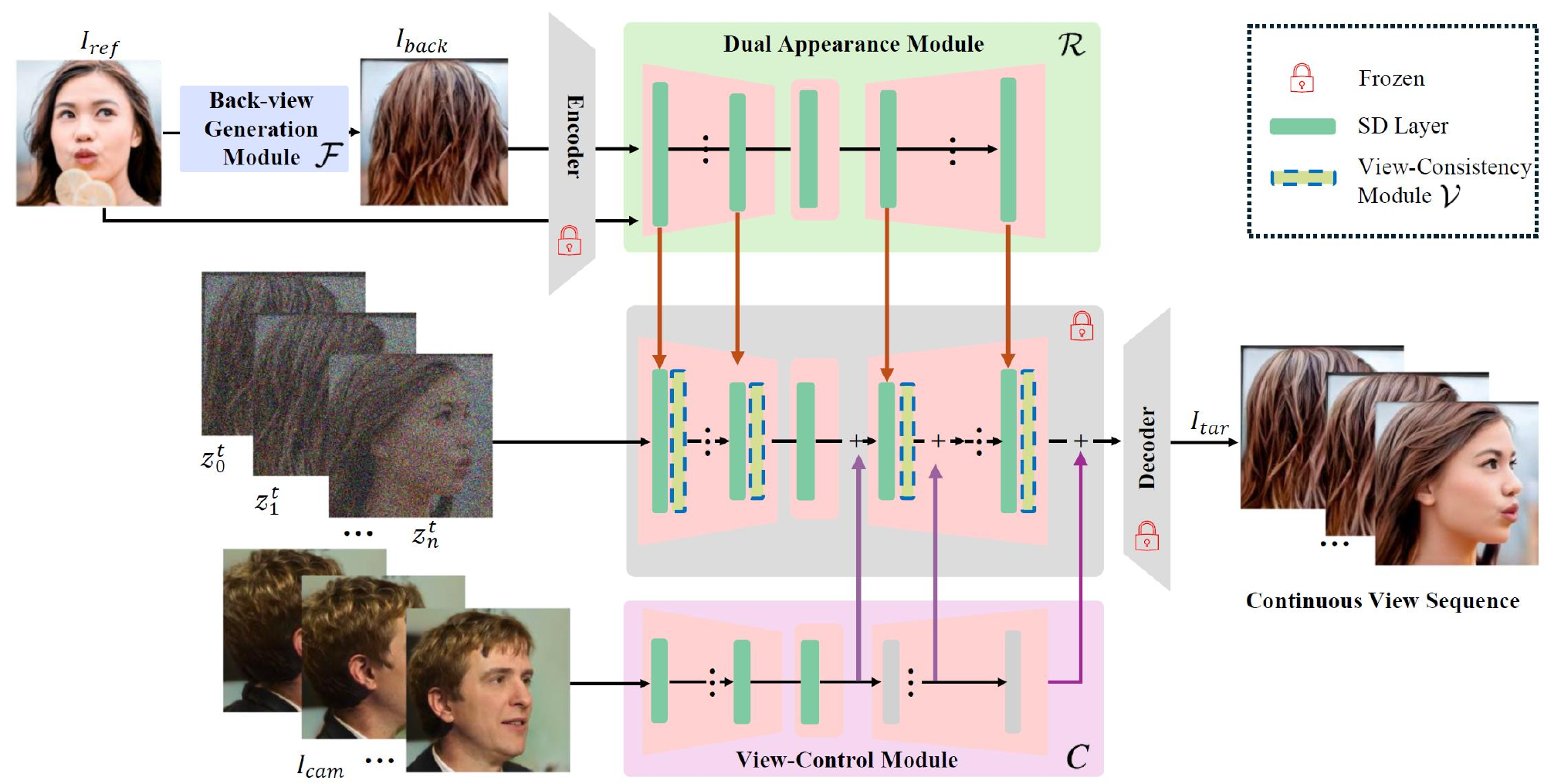}
\caption{
For the task of full-range 360-degree novel view synthesis, DiffPortrait360 employs a frozen pre-trained Latent Diffusion Model (LDM) as a rendering backbone and incorporates three auxiliary trainable modules for disentangled control of dual appearance \(\mathcal{R}\), camera control \(\mathcal{C}\), and U-Nets with view consistency \(\mathcal{V}\). Specifically, \(\mathcal{R}\) extracts appearance information from \(I_{\text{ref}}\) and \(I_{\text{back}}\), and \(\mathcal{C}\) derives the camera pose, which is rendered using an off-the-shelf 3D GAN. During training, we utilize a continuous sampling training strategy to better preserve the continuity of the camera trajectory. We enhance attention to continuity between frames to maintain the appearance information without changes due to turning angles. For inference, we employ our tailored back-view image generation network \(\mathcal{F}\) to generate a back-view image, enabling us to generate a 360-degree full range of camera trajectories using a single image portrait. Note that $z$ stands for latent space noise rather than image. }
\label{fig:pipeline}
\end{figure*}

%% file: sec/2_Related_Work.tex
\section{Related Work}
\input{figures/comparisons_style}
Numerous research methods and commercial solutions~\cite{pinscreenavatar,itseez3d} have been developed for reconstructing full 3D avatar heads from a single input photo, as opposed to using a multi-view capture approach~\cite{10.1145/1778765.1778777,10.1145/2070781.2024163,DBLP:journals/corr/abs-1808-00362, kirschstein2023nersemble, qian2023gaussianavatars}. These capabilities enable applications such as consumer-accessible immersive telepresence, personalized avatars for 3D games, and scalable approaches to content creation.
Despite their wide adoption, regression-based face modeling techniques from a single input image~\cite{paysan20093d, 10.1145/3130800.31310887, wu2019mvf, dib2021high, tsutsui2022novel, zhuang2022mofanerf, DBLP:journals/corr/abs-2106-11423} using parametric 3D Morphable Models (3DMM)~\cite{10.1145/311535.311556} are increasingly being complemented in many applications by neural rendering techniques, as surveyed in~\cite{DBLP:journals/corr/abs-2004-03805}.

\vspace{-3mm}
\paragraph{3D-Aware GAN-Based Full-Head Synthesis.}
Recent advancements on 3D-aware GANs~\cite{chan2021pi,gu2021stylenerf,or2021stylesdf,xue2022giraffe,Chan2022eg3d,epigraf,xu2021generative,deng2021gram,An_2023_CVPR,li2024spherehead}, often leveraging StyleGAN2~\cite{karras2020analyzing} as backbone, are capable of generating novel views from a single input photograph through fine-tuned GAN inversion while ensuring view-consistency using a volumetric neural radiance field representation based on tri-planes~\cite{Chan2022eg3d} or signed distance function, such as StyleSDF~\cite{or2021stylesdf}. 
While most of them can only generate frontal faces due their training data and their 3D representation, two recent methods have demonstrated how the generation of consistent 360-degree views are possible, namely, PanoHead~\cite{An_2023_CVPR} and SphereHead~\cite{li2024spherehead}. 
These methods generate a reasonable back head region using enhanced tri-plane representations by incorporating multiple depth layers or spherical coordinate system. 
However, their back-view synthesis often appear unnatural and mismatching the front view, especially when single-view inversion is applied. Most importantly, all of these methods are domain specific and only work on realistic human heads. They often fail for unknown accessories or complex hairstyles, and generating novel views for stylized avatars is not possible.

\paragraph{Diffusion Model Based 360-Degree View Synthesis.} 

Diffusion-based models~\cite{ho2020denoising, song2020denoising, song2020score}, particularly latent diffusion models (LDM)~\cite{rombach2022high}, have recently gained significant attention for their unmatched performance in generating highly diverse and high-fidelity images across various domains. A number of diffusion-based methods have been introduced for the task of novel view synthesis~\cite{Gu_2024_CVPR, liu2023one2345, liu2023zero1to3, shi2023zero123plus} for general objects, and several methods~\cite{shi2023zero123plus, Gu_2024_CVPR} demonstrate the effectiveness of fusing features of a reference image into self-attention blocks in the LDM UNets, facilitating high-quality novel view generation while preserved appearance context effectively. The notable work, ControlNet~\cite{zhang2023adding}, extends the LDM framework to controllable image generation with additive structural conditions from signals such as 3D-aware GAN-synthesized camera viewpoints or rendered 3DMM condition~\cite{prinzler2024joker}. The work, DiffPortrait3D~\cite{Gu_2024_CVPR} achieves the state-of-the-art portrait novel view synthesis results by seamlessly integrating the appearance and camera view attentions with pre-trained UNets. Despite the impressive results, especially for stylized input portraits, only frontal views are possible to generate, and unwanted local inconsistencies between the model from being converted into a 3D representation, such as textured Meshes, NeRFs or 3D Gaussian Splats, which is needed for free-viewpoint rendering. Another recent diffusion-based technique, called Portrait3D~\cite{hao2024portrait3d}, generates compelling views from all angles, achieving superior results to GAN-based approaches~\cite{An_2023_CVPR,li2024spherehead}  but is restricted to realistic heads only.
Our proposed solution is particularly effective in generating globally consistent and plausible back views of the head, while also ensuring locally view-consistent renderings for full $360^\circ$ style-omniscient portraits.

%% file: figures/comparisons_style.tex
\begin{figure*}
\centering
\includegraphics[width=1.0\linewidth]{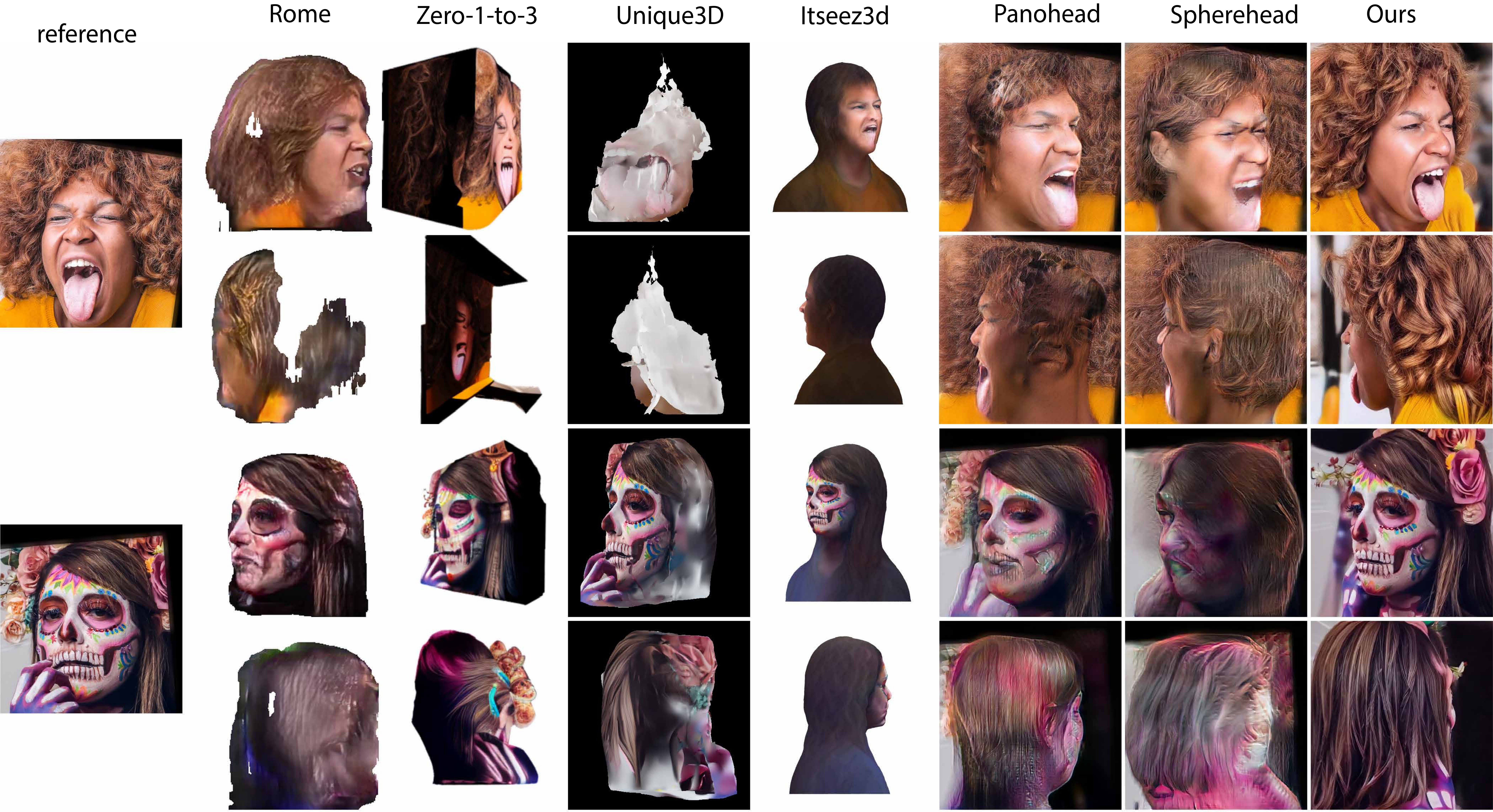}
\caption{Qualitative comparisons with existing methods on in the wild portraits. Compared to the baselines, our method shows superior generalization capability to novel view synthesis of wild portraits with unseen appearances, expressions, and styles, even without any reliance on fine-tuning.}
\label{fig:compare_style}
\end{figure*}

%% file: sec/3_Method.tex
\section{Method}

\input{figures/comparisons_real}
\input{figures/ablation_dual}
Given a single RGB portrait image with arbitrary style, \( I_{\text{ref}} \), our approach aims to synthesize a new image, \( I_{\text{tar}} \), from any camera perspective within a 360$^\circ$ range using an \( I_{\text{cam}} \) image. The novel view \( I_{\text{tar}} \) must maintain consistent appearance information with \( I_{\text{ref}} \), 
with the camera view controlled by \( I_{\text{cam}} \). It is crucial to note that this includes not only positions near \( I_{\text{ref}} \) but also areas with minimal overlap and previously unobserved regions from the given view.

In this work, DiffPortrait360 leverages powerful latent diffusion models, enabling disentangled control of appearance and camera views, as detailed in Section~\ref{method_pre}. 
Building upon this foundation, we propose a tailored dual-appearance module that incorporates additional appearance information from both the front and back of the head, \( I_{\text{back}} \) outlined in Section \ref{dual_appearance-method}. We introduce an additional ControlNet~\cite{zhang2023adding} based synthesizing network designed to generate a plausible back view in \( I_{\text{back}} \) from \( I_{\text{ref}}\), and to correct biases due to limited 3D head training data, as detailed in Section \ref{method_aux}. Finally, Section \ref{method_view} introduces a novel training strategy that leverages sequence priors to distill fine-grained structural information, ensuring consistency and smooth view transitions while maintaining high synthesis quality.


\subsection{Background}
\label{method_pre}
\paragraph{Latent Diffusion Model.} Diffusion models~\cite{ho2020denoising, song2020denoising, song2020score} are generative models designed to synthesize desired data samples from Gaussian noise $z_0$ by iteratively denoising across $T$ time steps. Latent diffusion models (LDMs)~\cite{rombach2022high} extend this framework by operating in an encoded latent space $\mathcal{E}(I)$ where $I$ is an input image, facilitated by a pre-trained auto-encoder $\mathcal{D} \circ \mathcal{E} (\cdot)$. Specifically,
the model is trained to learn the denoising process from $z_T \sim \mathcal{N}(0, 1)$ to $z_0 = \mathcal{E}(I)$ with the objective,
\begin{equation}
    L_{ldm} =\mathbb{E}_{z_0,t,\epsilon \sim \mathcal{N}(0,1)} \bigg[ \Big\lVert \epsilon-\epsilon_\theta \big(z_t,t\big) \Big\lVert_2^2 \bigg],
\end{equation} 
where $z_t$ is the noise map at timestep $t$ and $\epsilon_\theta$ is a trainable UNet equipped with layers of convolutions and self-/cross-attentions (TransBlock). Finally, the output image is calculated by mapping the denoised latent feature map back to the image domain $I = \mathcal{D}(z_0)$. Unlike traditional Text-to-Image (T2I) diffusion models, which rely on textual inputs to guide image synthesis, our approach directly extracts portrait appearance context and camera information from \(I_{\text{ref}}\) and \(I_{\text{cam}}\), respectively, without relying on text descriptions. Consequently, we use an empty prompt for text control and omit it from our formulations.
\paragraph{Diffusion-Based Novel View Synthesis.}
 Recent studies~\cite{lin2023consistent123,shi2023zero123plus, liu2023one2345, liu2023zero1to3} have leveraged pre-trained LDMs for novel view synthesis. These efforts largely employ latent diffusion as the backbone for novel view synthesis, incorporating modules for appearance injection and camera information control. While most of this research focuses on general objects, our work follows a specific ~\cite{Gu_2024_CVPR} designed for portrait novel view synthesis, which uses three main modules: 1) \textit{an Appearance Reference Module $\mathcal{R}$} that captures identity attributes and background from \( I_{\text{ref}} \), 2) a \textit{Control Module (C)} typically a ControlNet \cite{zhang2023adding}, that adjusts camera perspectives based on \( I_{\text{cam}} \) and 3) a \textit{View Consistency Module $\mathcal{V}$}, intervened within UNet blocks using temporal cross-attention between features of multiple consecutive views to ensure view consistency. We follow the same network structure but innovate on enabling 360-degree and consistent novel view synthesis that preserves input appearance coherence across extensive viewpoint shifts, encompassing elements such as expressions, hairstyles, and head shapes.
\input{figures/ablation_back}

\subsection{Dual Appearance Module}
\label{dual_appearance-method}
To ensure the preservation of appearance information characteristics, previous methods typically employ a ReferenceNet~\cite{cao2023masactrl} to inject the reference image and guide the denoising process with $I_{\text{ref}}$ in order to extract sufficient information from $I_{\text{ref}}$. The Appearance Control Module is trained to ensure meticulous transfer of referenced structure and appearance into the SD-UNet for the task of novel view synthesis of portrait views. While effective at reference control in front near face, such an appearance control scheme induces following problems: Firstly, the appearance content is largely provided by the image of $I_{\text{ref}}$. As the viewpoint shifts from the front of the face to the back of the head, the training dataset introduces randomly generated content such as different hairstyles, hats, and other features. Additionally, front face images contain only very limited information related to the back of the head, with very few pixels or even no pixels available for use. This leads to a significant misuse of most of the facial feature information from the reference image, causing leakage to the back of the head, which can be amplified to create a dual face or result in facial information appearing on the back of the head. Furthermore, these inaccuracies impact view control consistency, since the network, as a shortcut, tends to use other tailored modules, e.g., the view control module $\mathcal{C}$, to generate information when there is insufficient appearance information. As a result, system collapse occurs during the generation of larger views with only a single reference.

To mitigate this issue, we require an appearance reference network that could take information that entirely covers areas not observed by the front appearance while minimizing the overlap with the front face, which enables the diffusion network to access sufficient appearance information. This motivation leads us to design a dual-appearance module. Specifically, during training, we allow the appearance module to access two images, $I_{\text{ref}}$ and $I_{\text{back}}$. The rationale behind this modification is to reduce ambiguities caused by insufficient or erroneous information during viewpoint transitions, and to compel the network to autonomously decide which information from $I_{\text{ref}}$ and $I_{\text{back}}$ should be relied on more under different camera views. We note that in selecting $I_{\text{ref}}$ and $I_{\text{back}}$, views should overlap as little as possible but not be fixed, as this could affect the generalization ability of the input images. Consequently, in our training dataset, we deliberately choose pairs with minimal overlap as inputs for $I_{\text{ref}}$ and $I_{\text{back}}$, enabling our dual appearance module to leverage the information to its fullest extent. As a result, we observed consistent and satisfactory appearance in areas not covered by the reference images, and between the 360-degree area of $I_{\text{back}}$ and $I_{\text{ref}}$. This makes our model capable of generating consistent appearance information while changing the camera views(see Figure \ref{fig:ablation_dual}).
\input{figures/ablation_consistency}
\subsection{Back-View Reference Generation}
\label{method_aux}

Our trained dual appearance module, $\mathcal{R}$, significantly improves appearance consistency, especially during large changes in viewing angles. 
However, during inference, it is challenging to obtain a reasonable reference image context with a viewing angle drastically different from the input, such as obtaining a nearly opposite auxiliary image to the reference. 
Compromising to this difficulty will reduce the effectiveness of the disentangled appearance information determined by \( I_{\text{back}}\).

To address this issue, we developed a generative network capable of producing an \( I_{\text{back}} \) that has minimal overlap in appearance information with \( I_{\text{ref}} \) while maintaining a consistent style, reasonable head shape, and hairstyle. This motivates the use of ControlNet~\cite{zhang2023adding} and the original \( I_{\text{ref}} \) to generate a decoupled \( I_{\text{back}} \) as our extra appearance condition input. This approach is practical for real-world applications, allowing the generation of additional appearance information without needing external inputs.

However, this generation model can exhibit a strong bias when using only realistic data, which is unsuitable for producing a broader range of artistic styles. To avoid an over-reliance on a training dataset that could lead to the generation of unreasonable \( I_{\text{back}} \), we supplemented our dataset with a large collection of cartoon and stylized task images of back views produced by \cite{fluxai} and Unique3D~\cite{wu2024unique3d}, adjusting the distribution of our dataset to lessen the bias towards realistic backdrops. Notably, even with limited camera views (only including front and back views), this is sufficient for the Generative Module to correct biases from real dataset training, enabling it to generate reasonable unobserved head regions consistent with the input portrait style.
As demonstrated in our experiments, our trained $\mathcal{F}$, with a limited dataset of cartoon front and back faces, effectively resolves the challenge of sourcing additional appearance \( I_{\text{back}} \) in practical applications and generalizes well to unseen portrait styles. (see Figure \ref{fig:ablation_back}).

\subsection{View-Consistency Module} 
\label{method_view}
Given an arbitrary view portrait image \(I_{\text{ref}}\), our trained module has already been capable of generating a full 360-degree range of novel view syntheses \(I_{\text{tar}}\) via above designs. Nevertheless, during inference, we occasionally observe our module failing to capture nuanced view consistency compared to 3D-aware GANs, such as appearance flickering and choppy transitions between views. These unwanted artifacts significantly impact the effectiveness of novel view synthesis in downstream applications, such as 3D reconstruction.

Reviewing our above module, the view consistency module \(\mathcal{V}\) is primarily managed by a temporal transformer~\cite{guo2023animatediff}, integrated with 3D-aware noise~\cite{Gu_2024_CVPR}. However, given the heritage motion prior from~\cite{guo2023animatediff}, we note that view consistency is trained by focusing on cross-view attentions with randomly shuffled views. We argue this training approach does not fully utilize the advantages of the motion prior, which may lead to the observed choppy transitions and minor appearance changes, ultimately resulting in the loss of detail and collapse of downstream tasks.

To alleviate this issue, we have revised the training approach of the temporal transformer to include data from continuous view transitions. 
Instead of randomly sampling sparse views from a sparse multi-view dataset, we controlled the 3D-aware GAN~\cite{An_2023_CVPR} to generate consistently sampled sequential view changes. In contrast to random view cross-attention, our sequential training scheme effectively instructs the view consistency module $\mathcal{V}$ to maximize the benefits derived from the pretrained motion prior. Even when trained on only a limited synthetic dataset, our sequential training scheme is sufficient for $\mathcal{V}$ to decipher higher multi-view consistency accuracy and smoother view-changing results, enabling the synthesis process to produce smoother results that better fit downstream applications.

%% file: figures/comparisons_real.tex
\begin{figure*}
\centering
\includegraphics[width=1.0\linewidth]{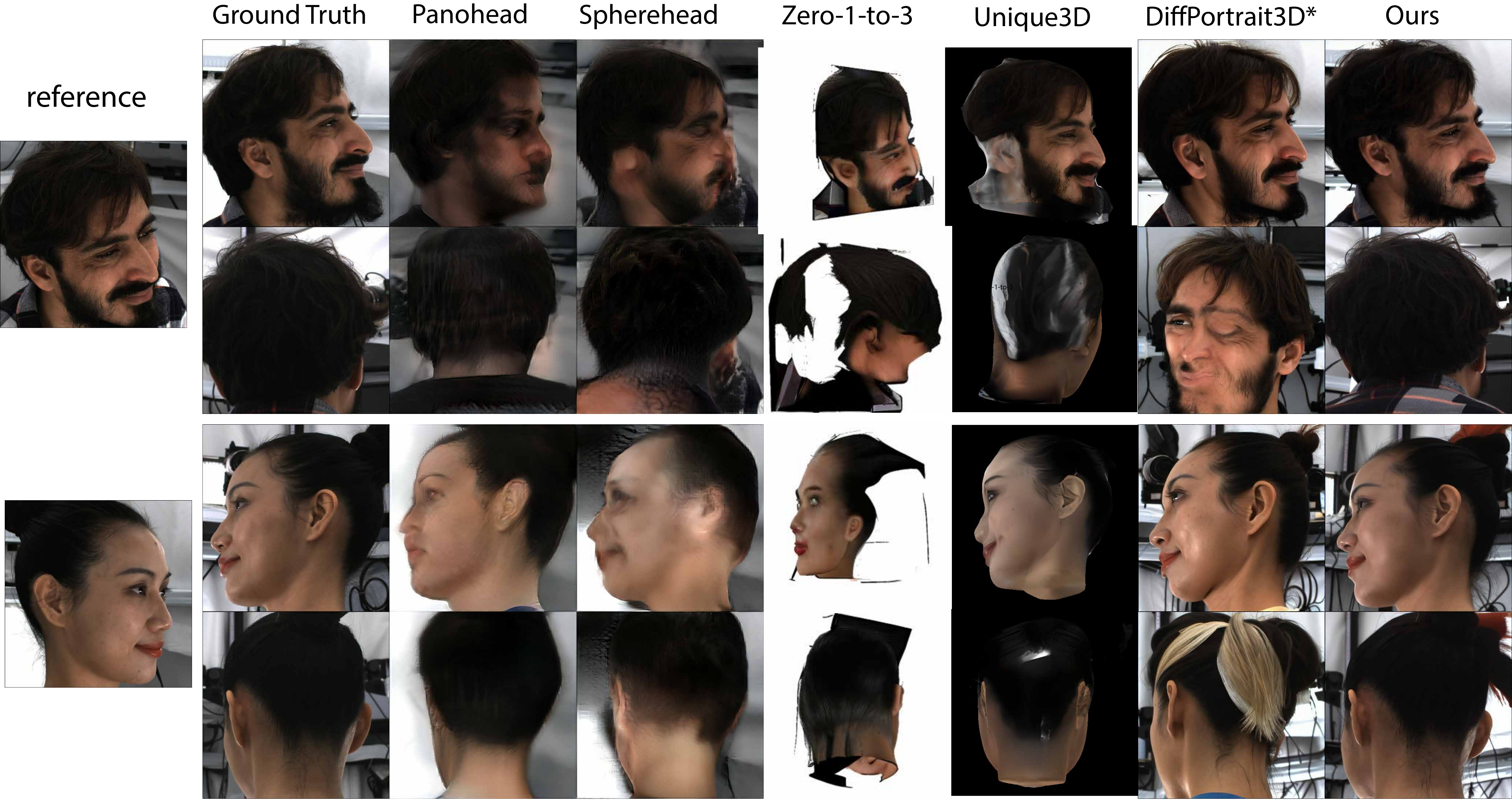}
\caption{Qualitative comparisons of novel view synterhsis on RenderMe360~\cite{pan2024renderme}. Our method achieves effective appearance control for novel synthesis under substantial change of camera view for synthesis.}
\label{fig:compare_real}
\end{figure*}

%% file: figures/ablation_dual.tex
 \begin{figure}
    \centering
    \includegraphics[width=0.97\linewidth]{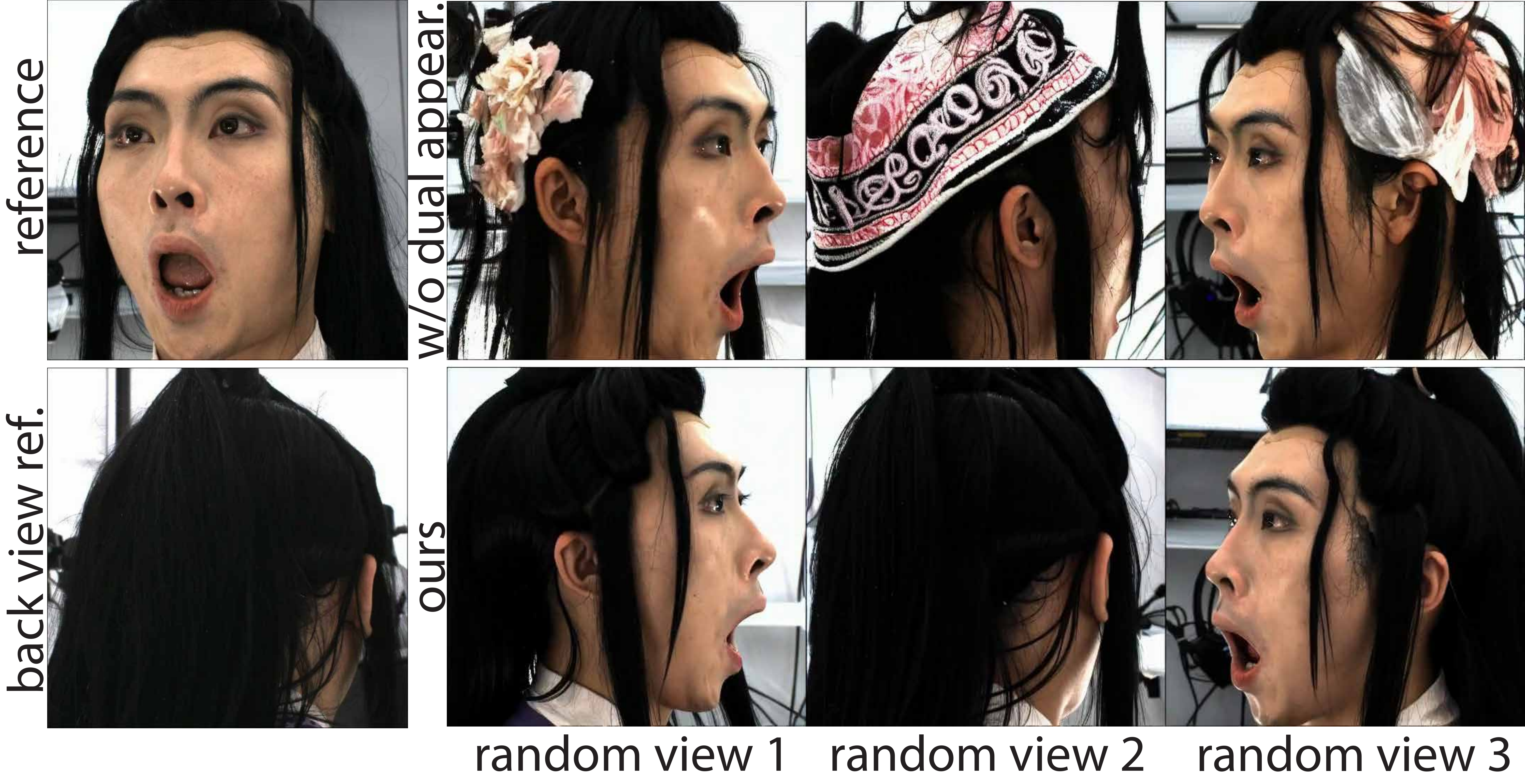}
    \centering
    \vspace{-3mm}
    \caption[Caption for LOF]{\textbf{Ablation Study on Dual Appearance Control.} }  
    \label{fig:ablation_dual}
\end{figure}

%% file: figures/ablation_back.tex
\begin{figure}
	\centering
	\includegraphics[width=0.97\linewidth]{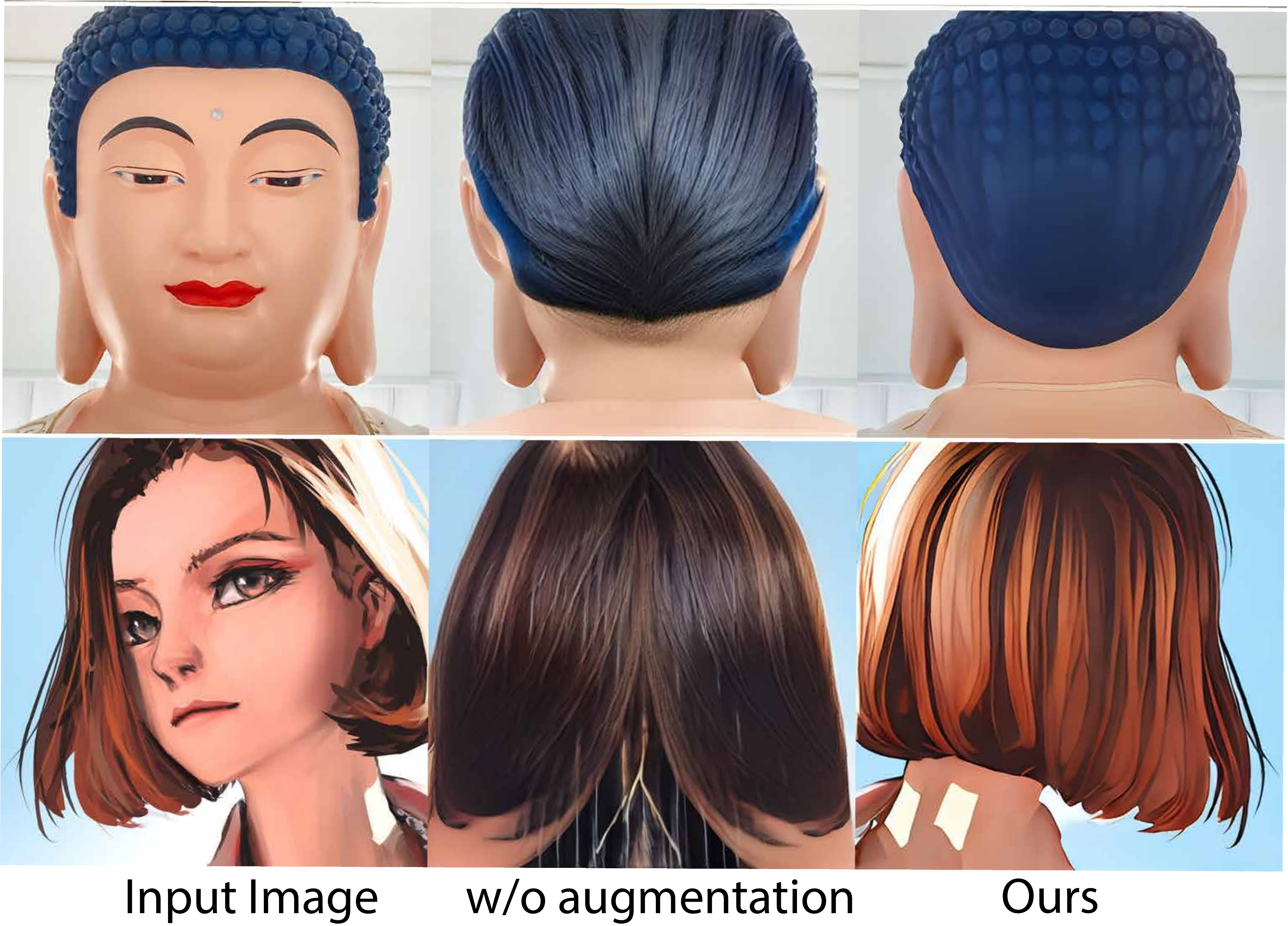}
	\centering
        \vspace{-3mm}
	\caption[Caption for LOF]{\textbf{Ablation on Back-view Generation.} }  
    \label{fig:ablation_back}
\end{figure}

%% file: figures/ablation_consistency.tex
 \begin{figure}
    \centering
    \includegraphics[width=0.97\linewidth]{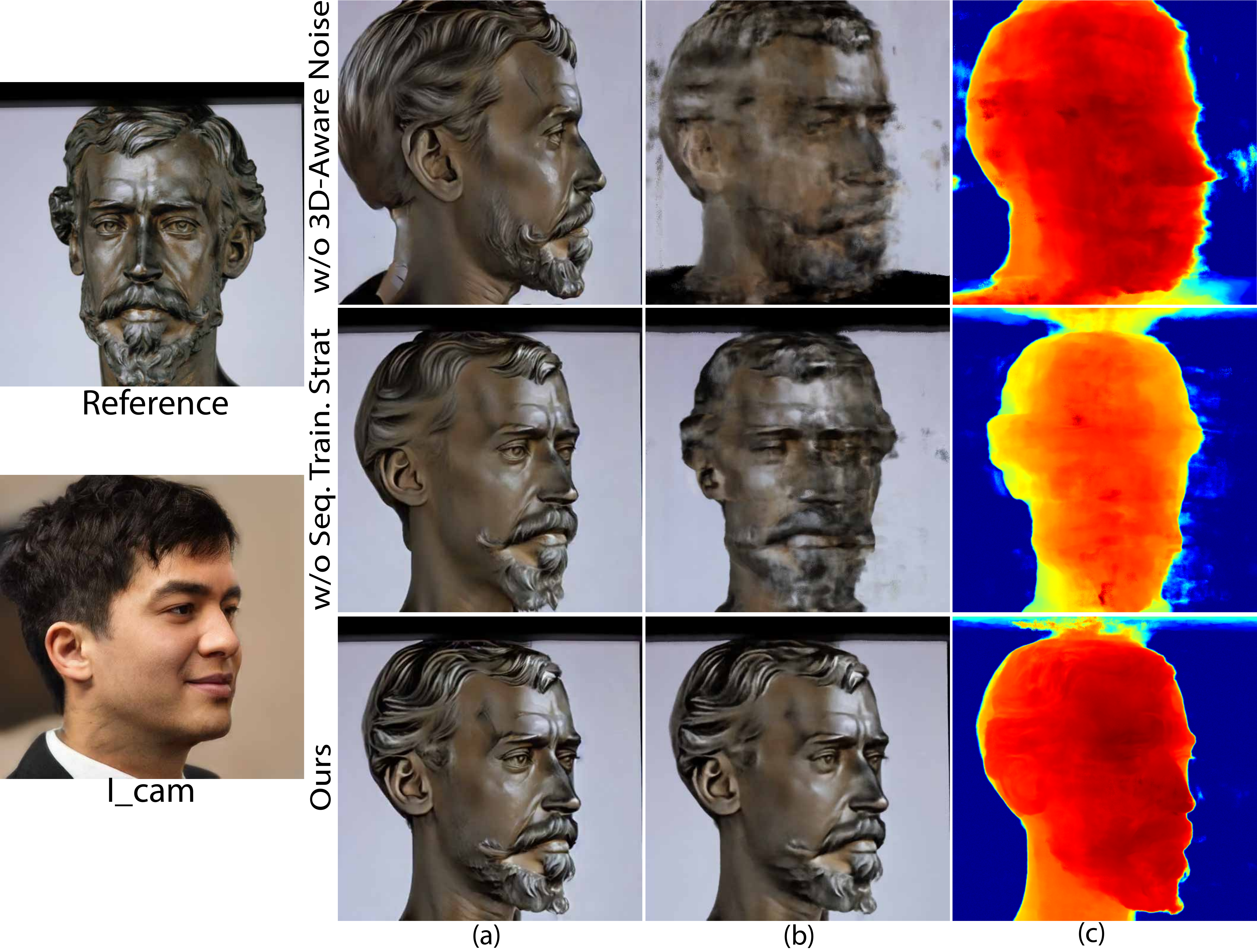}
    \centering
    \caption[Caption for LOF]{\textbf{Ablation Study on View-consistency Control.} (a) Generated novel view. (b-c) Rendering and depth of fitted NeRFs.}  
    \label{fig:ablation_consis}
\end{figure}

%% file: sec/4_Experiment.tex
\section{Experiments}


\subsection{Datasets}

We train our model using a hybrid dataset comprised of photo-realistic full-head multi-view images dataset RenderMe360~\cite{pan2024renderme} and synthetic ones by PanoHead~\cite{An_2023_CVPR} and SphereHead~\cite{li2024spherehead}. RenderMe360 consists of images from 500 individuals, each captured with 12 different expressions and an extended animated sequence from 60 multi-view camera positions. This setup offers a substantial amount of multi-view consistent appearance data for training. We manually separated the front and back camera information, selecting front and back images as inputs for dual-appearance control, while using the remaining camera views as training data. We randomly selected 1,000 frames of multi-view data from 150 subjects while using PanoHead and SphereHead to sample continuous views from 600 identities, specifically for training the sequential view consistency module due to the scarcity of sparse camera views of RenderMe360.
\input{tables/comparisons}
For the back-head reference image generation network, we collected 1,000 manually chosen, stylized front-and-back image pairs to augment the back-head image generation network. High-quality stylized front images were generated using a state-of-the-art tool~\cite{fluxai}; we then fed the images to Unique3D~\cite{wu2024unique3d} to generate content consistent back head. It’s important to note that this highly stylized dataset, containing only front and back views, was used solely as data augmentation to reduce bias in the back-view generation and was not included in the training of \(\mathcal{R}\).


\subsection{Comparisons}

\paragraph{Comparisons on Stylized Portraits.}
To demonstrate the generalization capability of our approach on stylized input portraits, we conduct a qualitative comparison with existing methods, as shown in \cref{fig:compare_style}. We also include Rome~\cite{khakhulin2022realistic} and Itseez3D\footnote{https://itseez3d.com/}. 
Thanks to our back-view generator and view-consistency module, our method exhibits superior generalization to diverse styles. 
We achieve high-quality details and continuous views across the full range of head poses. 
In contrast, it is evident that other approaches struggle to reach this level of performance.
Specifically, Rome fails to generate reasonable back-view textures and structures, especially for inputs with thick hair and unusual accessories. Similarly, Zero-1-to-3~\cite{liu2023zero1to3} and Unique3D~\cite{wu2024unique3d} continue to struggle with bridging the domain gap between general objects and human heads. 
Itseez3D appears to operate with templates that preserve only the facial textures from the reference image. 
Additionally, both PanoHead~\cite{An_2023_CVPR} and SphereHead~\cite{li2024spherehead} perform poorly on stylized portraits.

\paragraph{Comparisons on Real Portraits.}
We then compare the novel view synthesis results of our method with existing approaches on the test split of RenderMe360~\cite{pan2024renderme}, as illustrated in \cref{fig:compare_real} and \cref{tab:comparisons}. 
We additionally incorporate DiffPortrait3D*~\cite{Gu_2024_CVPR}, which has been finetuned using our full-head datasets, as the original model was limited to generating frontal views using the provided checkpoints. 
As shown in the qualitative comparisons in \cref{fig:compare_real}, the GAN-based methods, PanoHead and SphereHead, struggle to generate realistic details that align with the reference images. Their outputs fail to preserve identity due to the limitations of the PTI~\cite{roich2022pivotal}.
Zero-1-to-3 and Unique3D are trained on datasets of general synthetic 3D objects, and without specific domain knowledge, they cannot consistently produce reasonable novel views. Their results are often riddled with artifacts unrelated to heads.
Although we fine-tuned DiffPortrait3D* on our full-head datasets, it still struggles to generate consistent novel views at the back, resulting in significant discrepancies from the reference images. These examples show artifacts such as multiple faces and misaligned hair color.
In contrast, our specially designed method outperforms all others, demonstrating strong consistency across significantly different viewpoints. This conclusion is further supported by the quantitative comparisons detailed in \cref{tab:comparisons}. While existing methods reveal their shortcomings in evaluation statistics using perceptual metrics—especially for back views—our method performs nearly equally well for both frontal and back views.

\subsection{Ablation Study}

\paragraph{Back-view Generation.}
By assessing the multi-style frontal/back pairs augmented from outputs of Unique3D~\cite{wu2024unique3d}, we can clearly highlight their importance. As shown in \cref{fig:ablation_back}, when training the back-view generator using only real data pairs, the generated views align with the distribution of actual hair. In contrast, any incorrect back-view generated due to biases can severely impact the consistency of the subsequent full head view synthesis.

\vspace{-3mm}
\paragraph{Dual Appearance Control.}
As illustrated in the first row of \cref{fig:ablation_dual}, the generated near views (random view 1 and 2) can exhibit significant inconsistencies without leveraging our dual appearance control. Additionally, for the nearly opposite views 2 and 3, the generated hair accessories do not match, which occurs due to the limited reference information, as the target view has only a small overlap with the provided image. 
With the implementation of our dual control, enhanced by the generated back view, our method effectively addresses these issues, achieving excellent consistency in both textures and the underlying 3D shape, as demonstrated in the second row of \cref{fig:ablation_dual}.

\paragraph{View-Consistency Control.}
To assess the effectiveness of our strategy for view-consistency control, we conduct an ablation study with three configurations: 1) without 3D-aware noise, 2) without the sequential training strategy, and 3) our full method. 
In addition to comparing the synthesized views from diffusion, we also fit NeRF~\cite{park2021hypernerf} on 36 generated views that are evenly distributed around the head to thoroughly evaluate multi-view consistency. 
As shown in \cref{fig:ablation_consis}, the absence of 3D-aware noise results in a failure to achieve visual alignment between the diffusion output and the camera-control image. Without the sequential training approach, the method can achieve some visual alignment for certain views, as demonstrated in the second row of \cref{fig:ablation_consis}. However, when we fit the generated full-range 36 views into a NeRF using the approach from~\cite{park2021hypernerf}, the textures in rendered view (b) from the NeRF (with the same camera as view (a)) become significantly messy due to the discrepancies across multiple views.
In contrast, our complete method maintains good consistency among the generated views, which supports decent 3D fitting and shows ability to provide plausible depth. 


%% file: tables/comparisons.tex
\begin{table*}
    \centering
    \small
    \begin{tabular}{l|ccccc|cccc}
         \multirow{2}{*}{Method} & \multicolumn{5}{c|}{Frontal View} & \multicolumn{4}{c}{Back View} \\
         & PSNR $\uparrow$ & SSIM $\uparrow$ & LPIPS $\downarrow$ & CSIM $\uparrow$ & FID $\downarrow$ & PSNR $\uparrow$ & SSIM $\uparrow$ & LPIPS $\downarrow$ & FID $\downarrow$\\
         \hline
         PanoHead + PTI & 28.35 & 0.505 & 0.495 & 0.471 & 98.93 & 28.39 & 0.450 & 0.432 & 169.52\\
         SphereHead + PTI & 28.62 & 0.513 & 0.432 & 0.556 & 69.41 & 28.63 & 0.463 & 0.413 & 106.21\\
         \hline
         Zero123 & - & - & - & 0.431 & 152.03 & - & - & - & 216.88\\
         Unique3D & - & - & - & 0.483 & 195.33 & - & - & - & 285.75\\
         \hline
         DiffPortrait3D$^*$ & 28.96 & 0.509 & 0.425 & 0.709 & 49.02 & 28.47 & 0.373 & 0.446 & 91.37\\
         \hline
         Ours & \textbf{29.44} & \textbf{0.578} & \textbf{0.384} & \textbf{0.746} & \textbf{35.34} & \textbf{30.92} & \textbf{0.648} & \textbf{0.313} & \textbf{39.40} \\
    \end{tabular}
    \caption{Quantitative comparisons of novel view synthesis on RenderMe360~\cite{pan2024renderme} with perceptual metrics PSNR, SSIM~\cite{wang2004image}, LPIPS~\cite{zhang2018unreasonable}, FID~\cite{heusel2017gans} and an additional face recognition metric CSIM~\cite{zakharov2019few} for frontal views.}
    \label{tab:comparisons}
\end{table*}

%% file: sec/5_Conclusion.tex
\section{Discussion}

We have presented the \textit{first} framework for generating consistent, full 360° views from a single portrait, accommodating photorealistic and stylized humans, as well as anthropomorphic animals or statues. Our method is \textit{robust}, handling particularly difficult input portraits with \textit{complex accessories, make-ups, hairstyles, expressive faces}, and challenging \textit{lighting conditions}. We demonstrate that our back-of-head generation approach, combined with our novel dual appearance module, is essential for achieving globally consistent appearance generation full head.

\vspace{-3mm}
\section{Acknowledgment} This work is supported by the Metaverse Center Grant from the MBZUAI Research Office. We thank Egor Zakharov, Zhenhui Lin, Maksat Kengeskanov, and Yiming Chen for the early discussions, helpful suggestions, and feedback.


%% file: sec/X_suppl.tex
\clearpage
\setcounter{page}{1}
\maketitlesupplementary
\appendix

We provide additional implementation details in Section \ref{Implementation details}, highlight more comparisons and results in Section \ref{more reuslts}, and provide additional limitations in Section \ref{limitation and future work}.

\section{Implementation Details}
\label{Implementation details}
\subsection{Back-view Generation Module}
 We illustrate the framework of our back-view generation module $\mathcal{F}$ in~\cref{fig:pipeline} along with its training data examples in~\cref{fig:ablation_back_sel}, which is used to enhance the generative capabilities of $\mathcal{F}$. Specifically, we employ Stable Diffusion SD1.5 \cite{rombach2022high} as the backbone. A reference network \cite{cao2023masactrl} is utilized to inject information from the front face, and camera control is managed by ControlNet \cite{zhang2023adding}. In practice, we set a fixed camera view of the back head. In particular, we uniformly generate a fixed view that captures the maximum back appearance information, allowing the entire back-view generation module to focus on back-view appearance generation.

For the stylized augmentation dataset used to train the back-view generation module $\mathcal{F}$, we generate 2,000 subjects that are various stylized front portraits using \cite{fluxai} and further produce ground truth back-view by \cite{wu2024unique3d}. All data were processed with a cropped resolution of $512\times512$. Some of the lower-quality data were filtered out using \cite{deng2020disentangled}.
\input{figures/supply/pipeline}

 \input{figures/supply/supp_back}

\subsection{Training and Evaluations} 
Our model keeps the weights of the original Stable Diffusion model frozen during training. The training was conducted in stages, where we sequentially integrate and train modules $\mathcal{R}$, $\mathcal{C}$, and $\mathcal{V}$ mentioned in main paper Figure 2. All training were conducted on 6 NVIDIA RTX A6000 ADA GPUs with a learning rate of $10^{-5}$; we performed 60,000 iterations with the enhanced dataset for dual appearance and an additional 60,000 iterations for the camera control stages and the sequential training stage each, using only the PanoHead data due to unachievable camera intrinsics of the back view and a limited number of sparse camera views. For test inference, we collect 200 challenged portraits from Midjourney and Pexels \cite{midjourney, pexels}, containing a wide variation in appearance, expression, camera perspective, and style. For comparison in RenderMe360\cite{pan2024renderme}, we use another unseen 500 multi-view image pairs with different expressions.

\subsection{Dataset Details}

Our dataset has 800 unique subjects: 150 from RenderMe360 and 600 from PanoHead/SphereHead. We use 150 from RenderMe360, excluding those with complex head accessories to avoid unwanted artifacts. RenderMe360 is not used for sequential training due to sparse camera views. For the back-head generator, we use 1,800 subjects: 150 from RenderMe360 (real-world), 650 from PanoHead/SphereHead (synthetic), and 1,000 from Unique3D (stylized).

\section{More Results}
\label{more reuslts}

\subsection{More Ablation Study}
\begin{figure}
    \centering
    \includegraphics[width=1.0\linewidth]{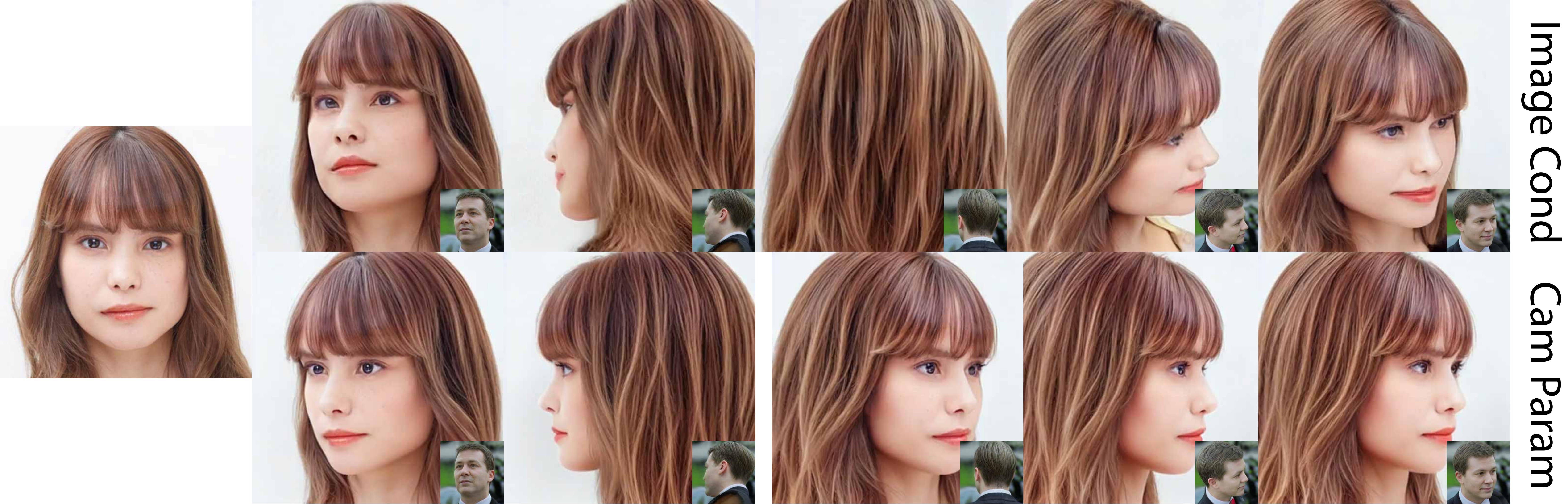}
    \caption[Caption for LOF]{Ablation study on camera conditioning.}
    \label{fig:cam_parm_ablation}
\end{figure}

As shown in~\cref{fig:cam_parm_ablation}, our experiments show image-based view control is more accurate than naive camera pose representation via text embedding. The effectiveness of alternative latent pose representations remains unclear, and explicit pose estimation requires extensive high-quality data, which we lack.
However, we compare both methods using the POSE metric in Tab. 1 of DiffPortrait3D [14].
(POSE$\downarrow$ Our Image Cond.: \textbf{0.0018}, Cam Param.: 0.0081).

\subsection{More Qualitative Comparisons} 

We conduct more qualitative comparisons with PanoHead~\cite{An_2023_CVPR}, SphereHead~\cite{li2024spherehead}, Unique3D~\cite{wu2024unique3d}, Rome~\cite{khakhulin2022realistic} and Itseez3D\footnote{https://itseez3d.com/} on stylized portraits in ~\cref{fig:compare_style_more}, where our method shows superior generalization capability on diverse styles. We also show additional qualitative comparisons with PanoHead~\cite{An_2023_CVPR}, SphereHead~\cite{li2024spherehead}, Zero-1-to-3~\cite{liu2023zero1to3}, Unique3D~\cite{wu2024unique3d} and DiffPortrait3D~\cite{Gu_2024_CVPR} on RenderMe360\cite{pan2024renderme} with paired ground truth results in ~\cref{fig:compare_real_more}.

\subsection{Comparing DiffPortrait3D on More Views} 

\input{figures/supply/moreview}
Please find more comparisons with DiffPortrait3D~\cite{Gu_2024_CVPR} on more views in~\cref{fig:moreview}.

\subsection{More Results}
Finally, more visual results of our method are introduced in~\cref{fig:more_results_0}, ~\cref{fig:more_results_1} and~\cref{fig:more_results_2}.

\input{figures/supply/more_comparisons_style}

\input{figures/supply/more_comparisons_real}

\input{figures/supply/more_results}

\section{Limitation and Future Work}
\label{limitation and future work}
\input{figures/supply/limitation}

Our experiments prove unlike the temporal fusion layer of DiffPortrait3D trained on random views, our approach using continuous improves local consistency which is crucial for 3D tasks. To this end, we also show that our emphasis on achieving view consistency is crucial for constructing a NeRF representation, enabling real-time free-viewpoint rendering from any camera position. While our method outperforms all state-of-the-art techniques, it still exhibits small inconsistencies for certain portraits. This is due to the inherent 2D nature of stable diffusion generation, which remains frozen during training to maintain stability. In the future, we plan to either incorporate geometric priors explicitly into the diffusion-based view synthesis process or extend our framework by directly encoding multi-view images into visual patches similar to, e.g., SORA\cite{videoworldsimulators2024} and explore the use of differentiable rendering techniques and efficient radiance field representations, such as 3D Gaussian Splats. Additionally, our method currently struggles with certain types of headgear, such as various hats and unseen hairstyles shown in~\cref{fig:limitation}, due to a biased distribution in our training data. We believe that additional data collection is likely to improve the performance. Finally, as our generated heads are currently static, future directions include making them animatable and relightable, and the cropping size of the head area should be expanded to accommodate scenarios involving longer hair.

%% file: figures/supply/pipeline.tex
\begin{figure}
\centering
\includegraphics[width=1.0\linewidth]{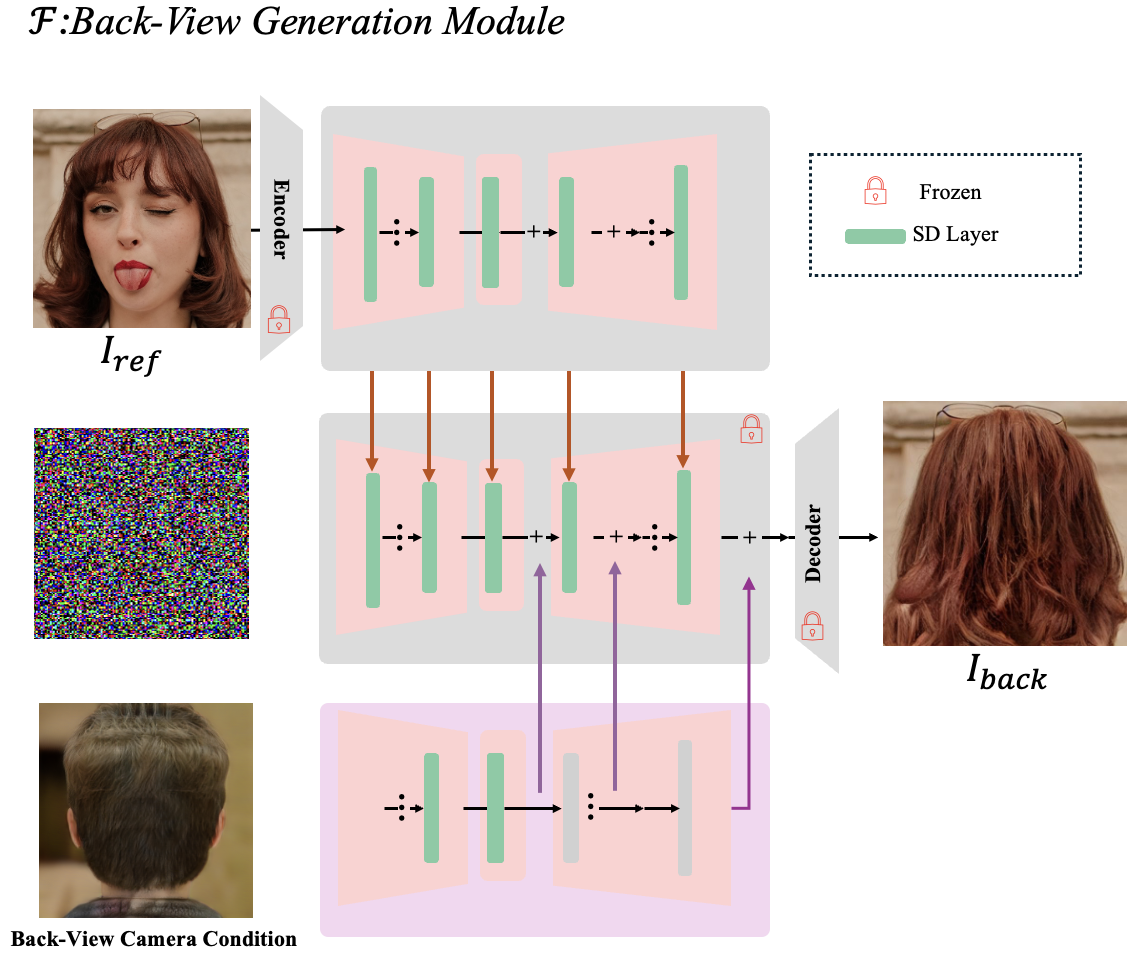}
\caption{Illustration of our back-view generation module $\mathcal{F}$. Given an arbitrary back-view camera condition and a reference image, we follow the methods of the reference module and ControlNet to decode a specific camera view. During inference, we set the back-view camera generation condition to 180 degrees to maximize the capture of appearance information from the back view.}
\label{fig:pipeline}
\end{figure}

%% file: figures/supply/supp_back.tex
\begin{figure}
	\centering
	\includegraphics[width=0.97\linewidth]{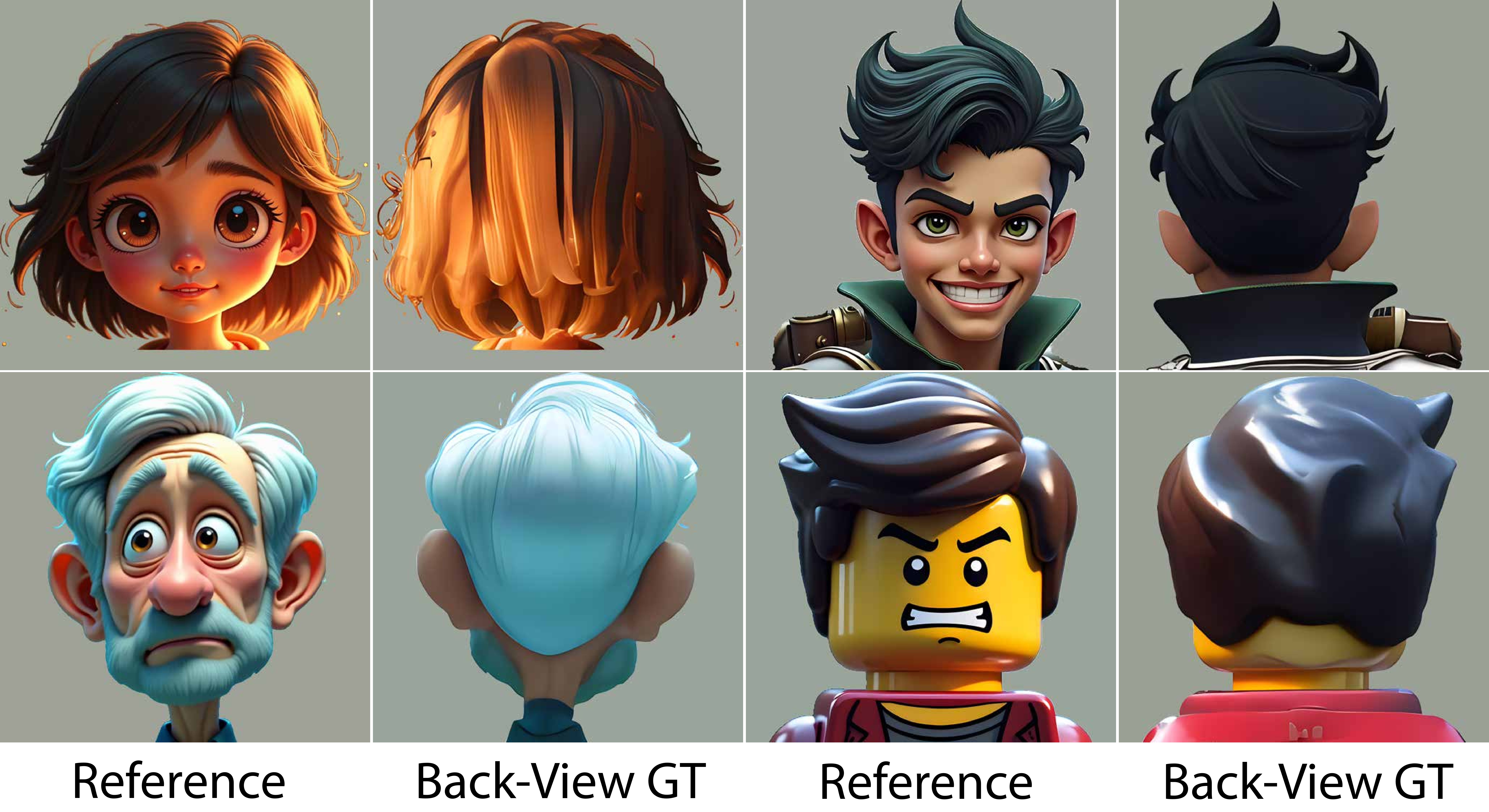}
	\centering
	\caption[Caption for LOF]{Examples in our stylized data augmentation which includes extensive generation of stylized back appearances. Compared to training back-view generation module $\mathcal{F}$ solely on real or synthetic networks, such augmentation helps our back-view generation module achieve greater generalizability.}
    \label{fig:ablation_back_sel}
\end{figure} 

%% file: figures/supply/moreview.tex
\begin{figure*}
\centering
\includegraphics[width=1.0\linewidth]{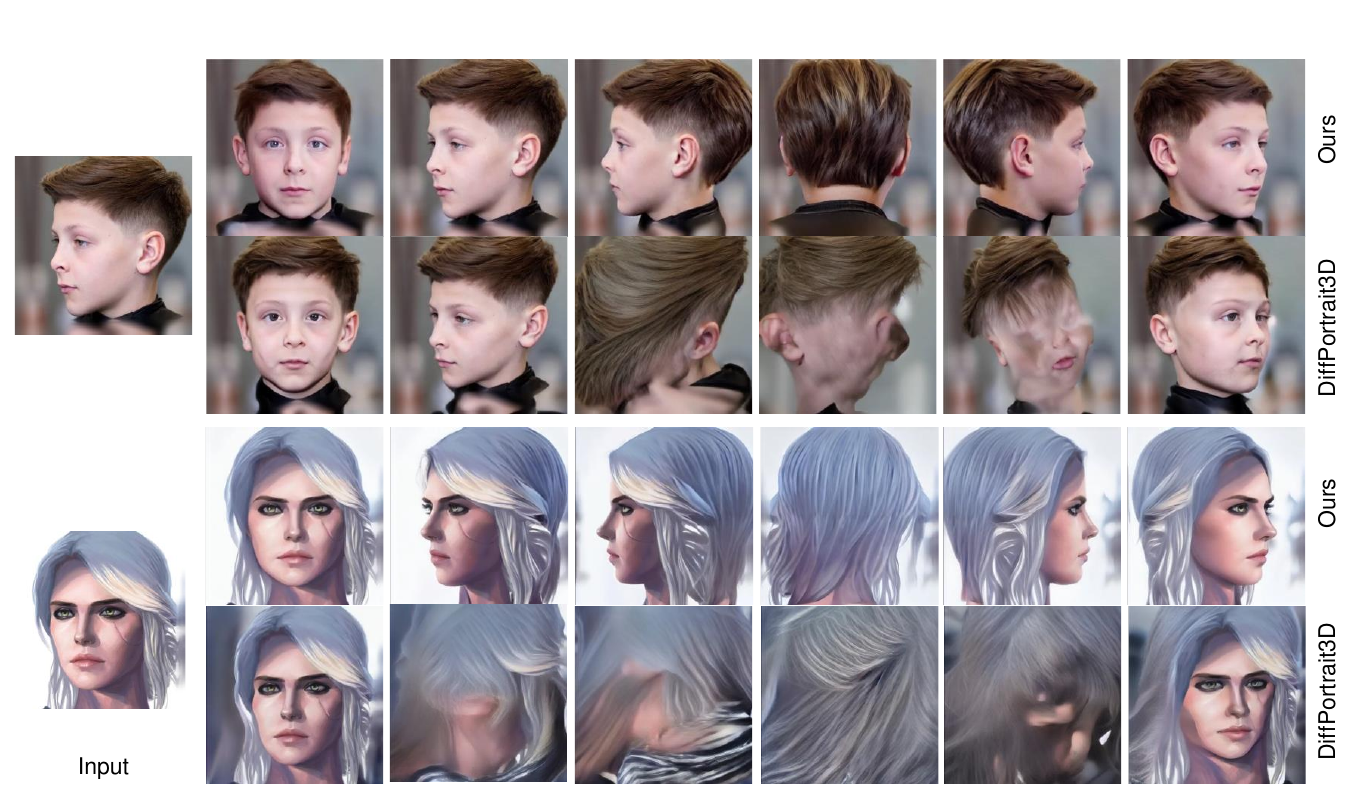}
\caption{More comparisons with DiffPortrait3D~\cite{Gu_2024_CVPR} on more views.}
\label{fig:moreview}
\end{figure*}

\begin{figure*}
\centering
\includegraphics[width=1.0\linewidth]{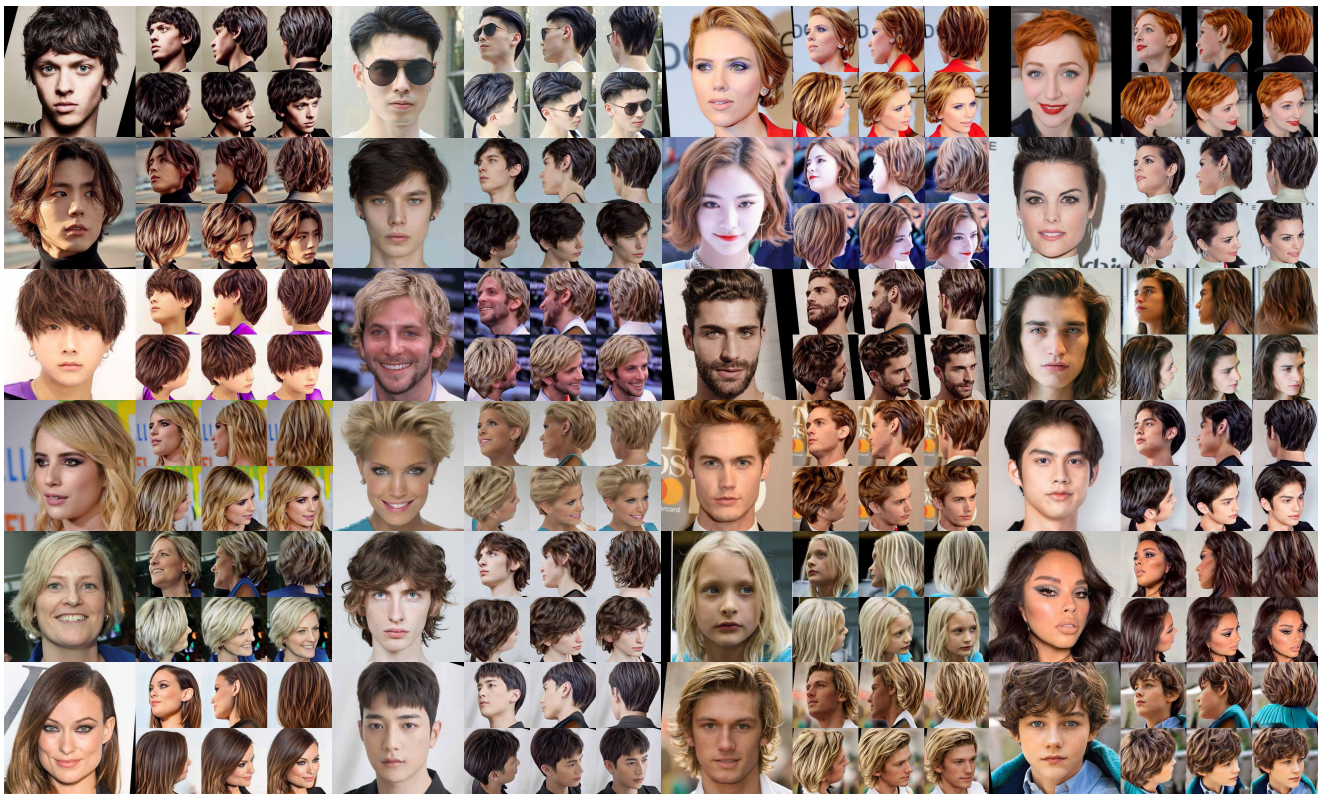}
\caption{More real-world results.}
\label{fig:realworldresults}
\end{figure*}

%% file: figures/supply/more_comparisons_style.tex
\begin{figure*}
\centering
\includegraphics[width=0.85\linewidth]{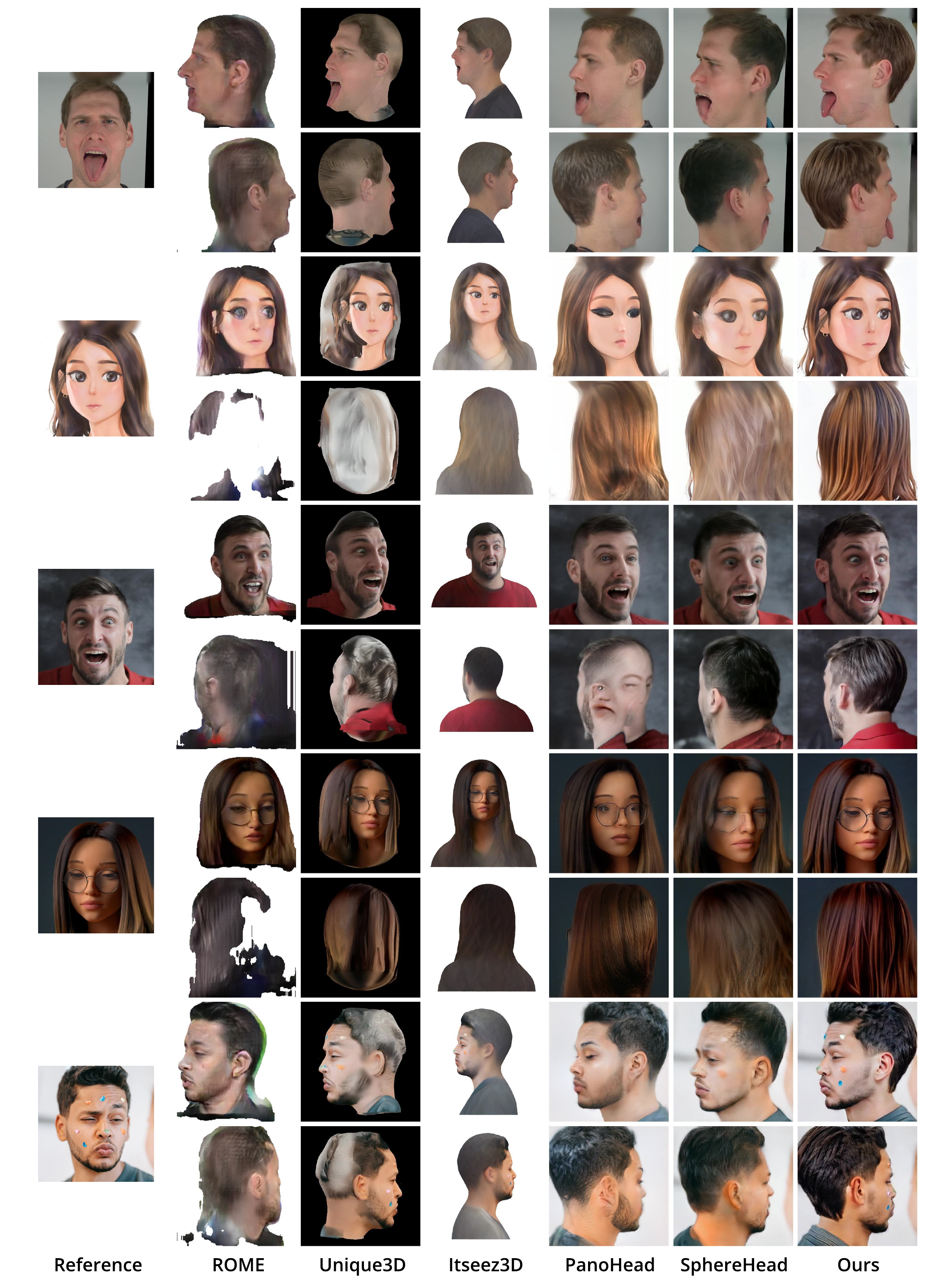}
\caption{More qualitative comparisons with existing methods on in the stylized portraits. Our method shows superior generalization capability to novel view synthesis of wild portraits with unseen appearances, expressions, and styles, even without any reliance on fine-tuning.}
\label{fig:compare_style_more}
\end{figure*}

%% file: figures/supply/more_comparisons_real.tex
\begin{figure*}
\centering
\includegraphics[width=1.0\linewidth]{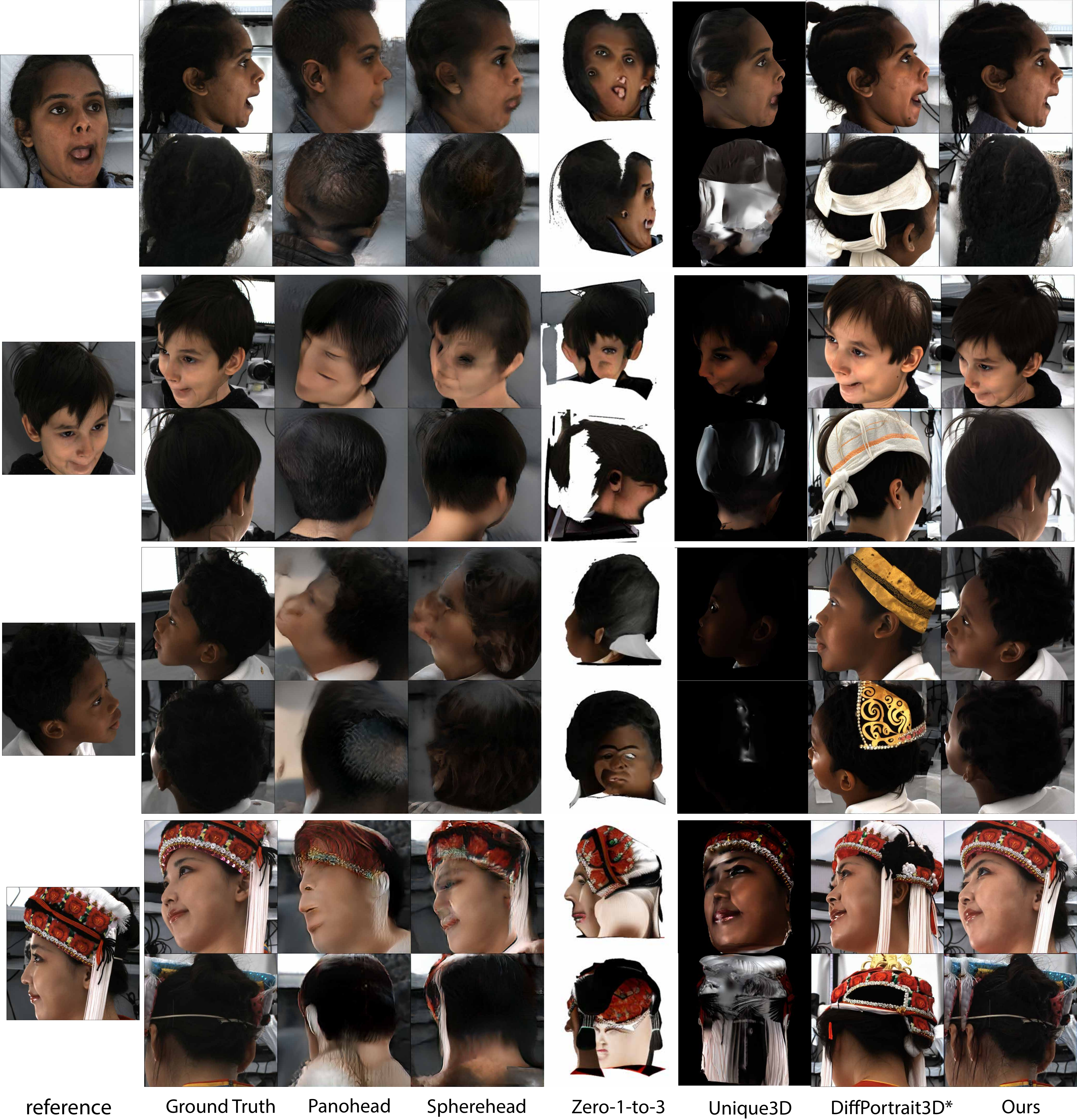}
\caption{More qualitative comparisons of novel view synthesis on RenderMe360~\cite{pan2024renderme}. Our method achieves effective appearance control for novel synthesis under substantial change of camera view for synthesis.}
\label{fig:compare_real_more}
\end{figure*}

%% file: figures/supply/more_results.tex
\begin{figure*}
\centering
\includegraphics[width=0.95\linewidth]{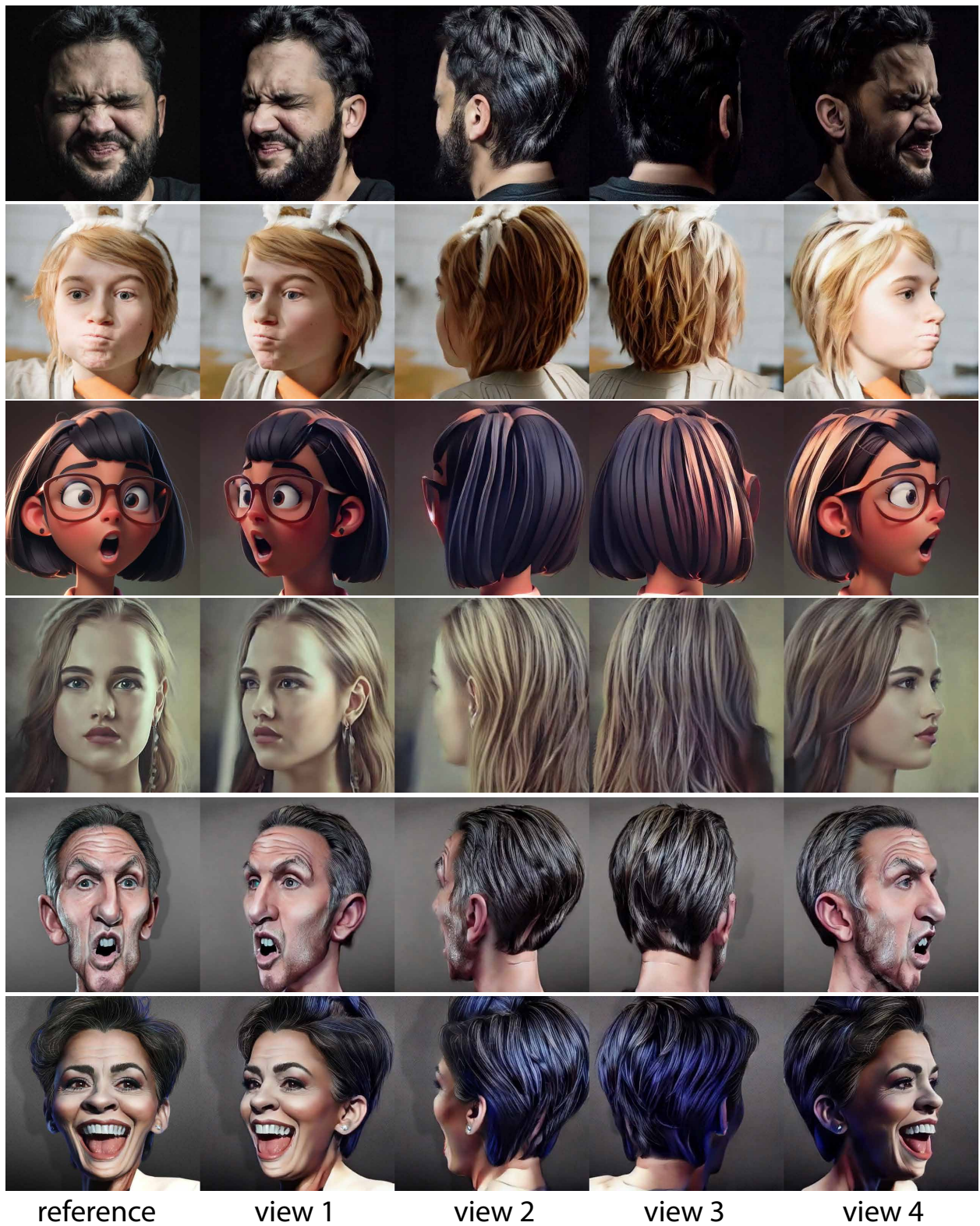}

\caption{More qualitative results of our method.}
\label{fig:more_results_0}
\end{figure*}

\begin{figure*}
\centering
\includegraphics[width=0.95\linewidth]{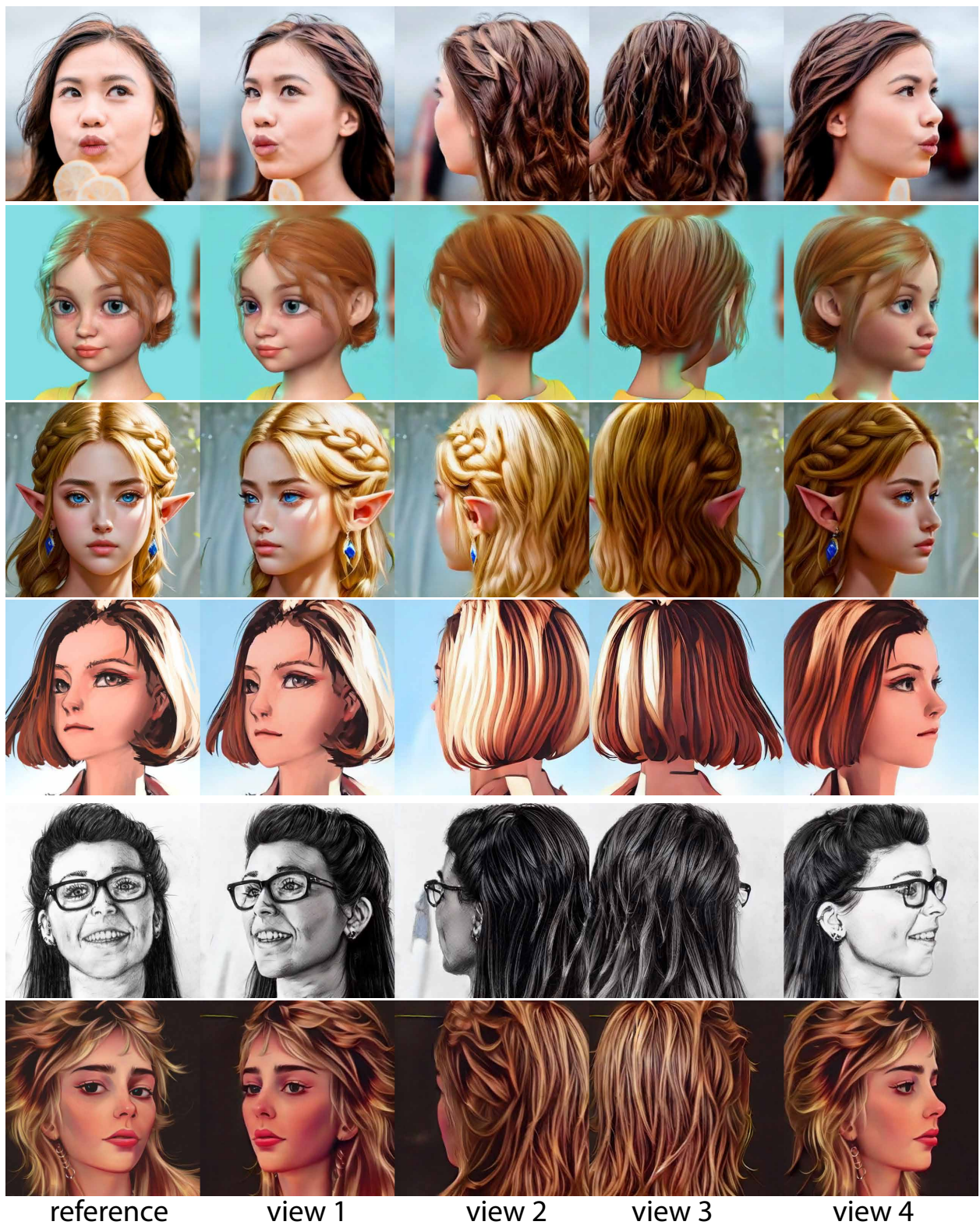}

\caption{More qualitative results of our method.}
\label{fig:more_results_1}
\end{figure*}

\begin{figure*}
\centering
\includegraphics[width=0.95\linewidth]{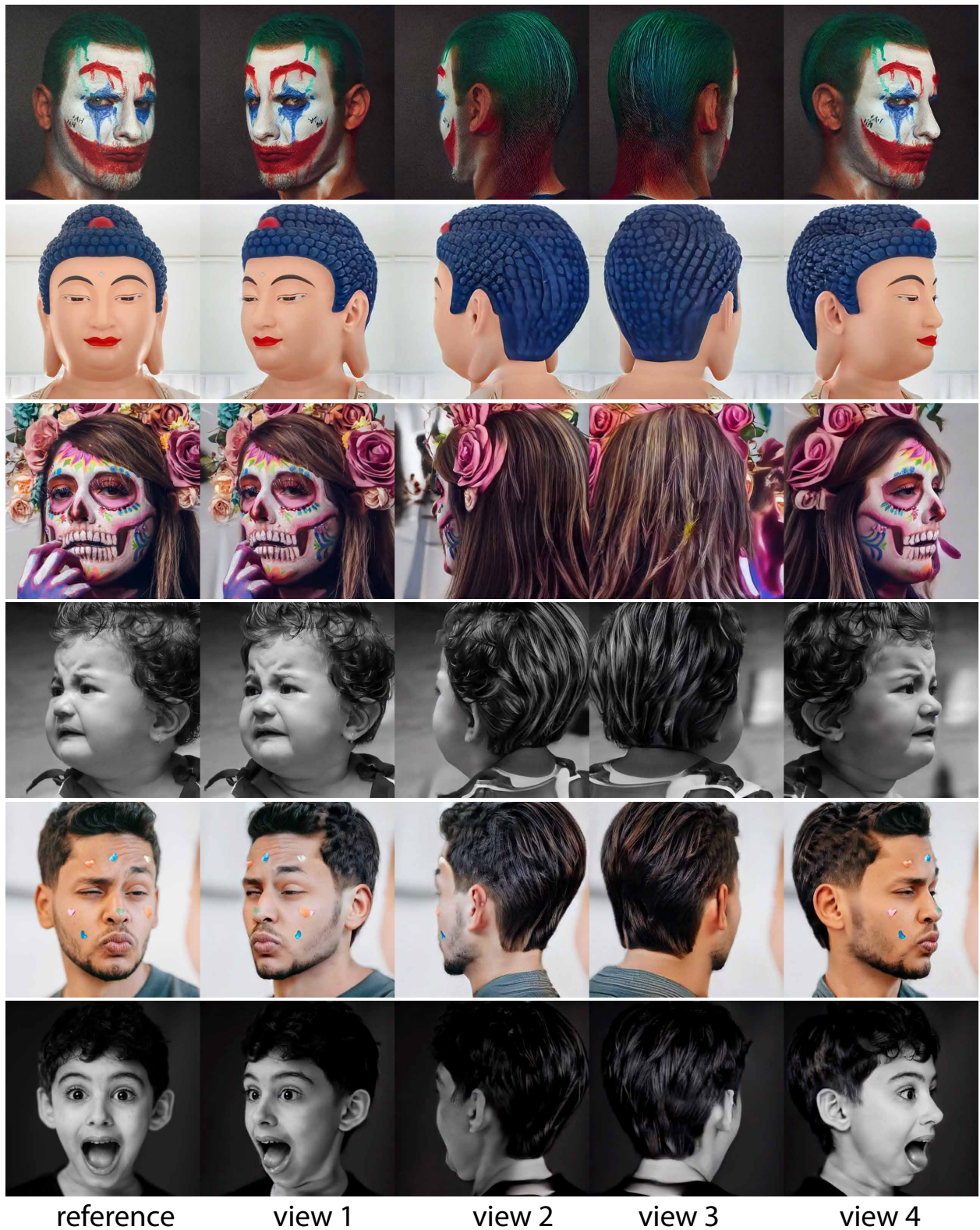}
\caption{More qualitative results of our method.}
\label{fig:more_results_2}
\end{figure*}

%% file: figures/supply/limitation.tex
\begin{figure*}
\centering
\includegraphics[width=1.0\linewidth]{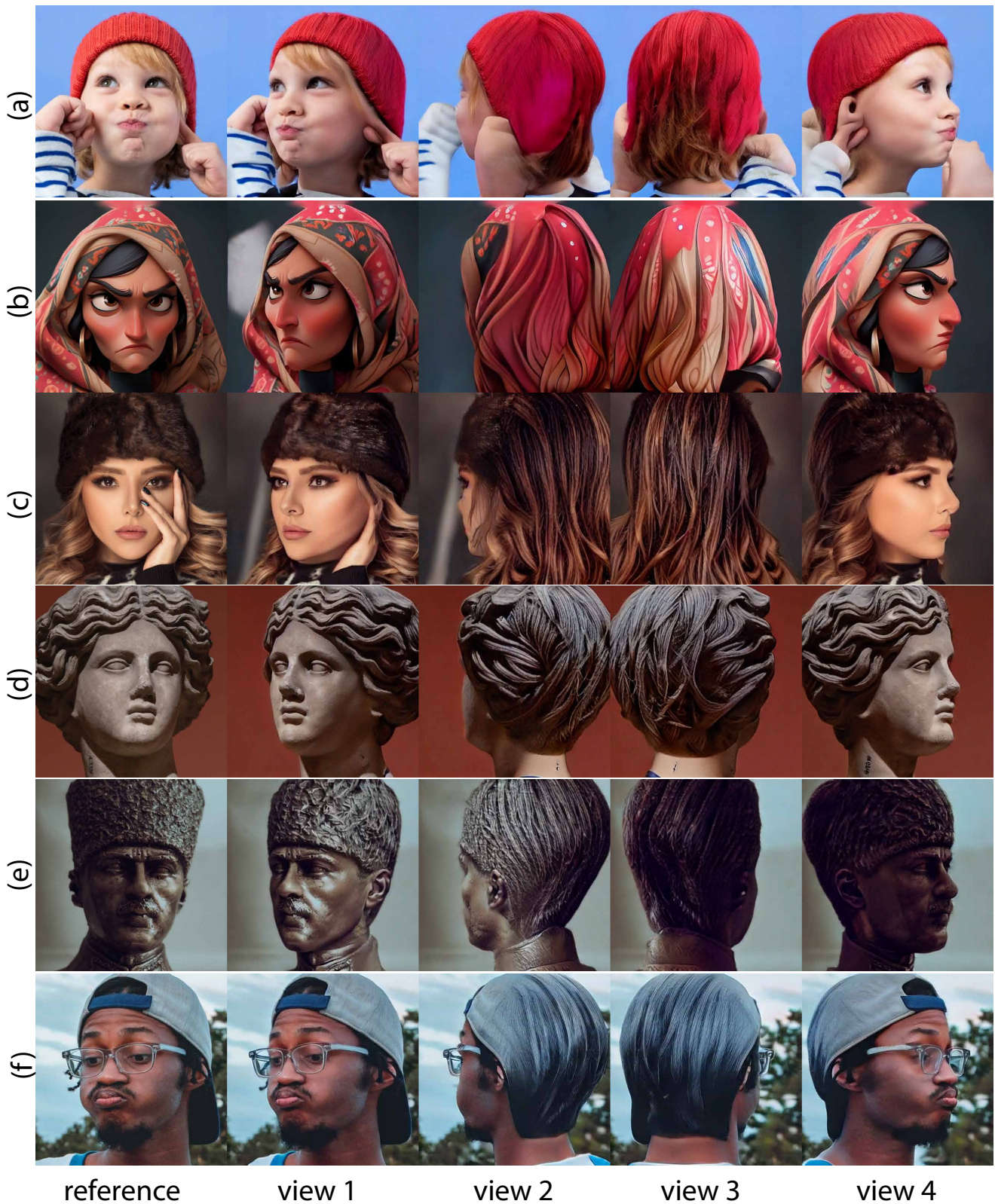}
\caption{Limitations of our method.}
\label{fig:limitation}
\end{figure*}